\documentclass{article}

\usepackage{microtype}
\usepackage{graphicx}
\usepackage{booktabs} 
\usepackage{url}            
\usepackage{array}
\usepackage{multirow}
\usepackage{amsfonts}       
\usepackage{nicefrac}       
\usepackage{todonotes}
\usepackage{siunitx}
\usepackage{etoolbox}
\usepackage[inline]{enumitem}
\usepackage{subcaption}
\usepackage{mathrsfs} 
\usepackage{mathtools} 
\usepackage{bm} 
\usepackage{amsmath} 
\usepackage{amssymb}

\usepackage{hyperref}

\DeclarePairedDelimiter\abs{\lvert}{\rvert}%
\newcommand{\p}[1]{\left(#1\right)}
\newcommand{\qb}[1]{ \left[#1 \right]}

\newcommand{\cb}[1]{\left \{ #1 \right \}}

\usepackage[accepted]{icml2019}

\icmltitlerunning{Graph Convolutional Gaussian Processes}

\begin{document}

\twocolumn[
\icmltitle{Graph Convolutional Gaussian Processes}



\icmlsetsymbol{equal}{*}

\begin{icmlauthorlist}
\icmlauthor{Ian Walker}{icl}
\icmlauthor{Ben Glocker}{icl}
\end{icmlauthorlist}

\icmlaffiliation{icl}{Department of Computing, Imperial College London, United Kingdom}

\icmlcorrespondingauthor{Ian Walker}{ian.walker14@imperial.ac.uk}

\icmlkeywords{Machine Learning, ICML}

\vskip 0.3in
]



\printAffiliationsAndNotice{}  

\begin{abstract}
We propose a novel Bayesian nonparametric method to learn translation-invariant relationships on non-Euclidean domains. The resulting graph convolutional Gaussian processes can be applied to problems in machine learning for which the input observations are functions with domains on general graphs. The structure of these models allows for high dimensional inputs while retaining expressibility, as is the case with convolutional neural networks. We present applications of graph convolutional Gaussian processes to images and triangular meshes, demonstrating their versatility and effectiveness, comparing favorably to existing methods, despite being relatively simple models.
\end{abstract}

\section{Introduction}

We present a new Gaussian process (GP) model called the graph convolutional GP (GCGP) model. This model learns translation-invariant relationships that mimic the behavior of a convolutional layer on images, but extends this property to general graphs. Since this is a GP model, the patch response function learned will be nonlinear and nonparametric, which may aid in expressibility and make more efficient use of small datasets. Additionally, since this is a Bayesian technique, it allows for the rigorous treatment of uncertainties for predictions, useful in many application domains, but difficult for existing convolutional models.

The main point of attempting to utilize general convolutions is to reduce the complexity of the model while maintaining expressibility. Indeed, standard GP models can suffer from the curse of dimensionality, making stable estimation difficult on high-dimensional inputs \citep{rasmussen2004gaussian}. This work seeks to create a convolutional model for general graphs, such as images, social networks, or 3D meshes, that focuses attention on local patches to substantially reduce the dimensionality of the input in a principled manner, inspired by recent work in geometric deep learning \citep{gdl}. In this way, GP models can be applied to high-dimensional inputs that live on non-Euclidean domains. To the best of our knowledge, this is the first work to propose \emph{spatial} graph convolutions with Gaussian processes.

We present applications of GCGPs to graphs which are Euclidean sampling grids (images) and further demonstrate the GCGP's performance when learning on non-Euclidean domains for classification tasks. We apply our method to triangular meshes and to an MNIST superpixel dataset, where each image is represented as a distinct graph.
While graph convolutional GPs are shallow, though wide, the results are promising for such relatively terse models, and indicate that GCGPs can provide a simple and effective foundation for more complex models in the future.

\section{Background}
A comprehensive overview of related graph kernel methods is given by \citet{vishwanathan2010graph} with some recent work by \citet{neumann2016propagation} proposing kernels for graphs with node features and other work employing deep learning \citep{zhang2018end,duvenaud2015convolutional}.

We focus on the task of classifying signals that live on general graphs, leaving regression tasks as a straightforward extension of the methodology. We further constrain our focus to domains which are the same size for each signal to be classified. Thus, our datasets will comprise of general graphs, which are composed of the same number of vertices. However, they do not necessarily have the same edge structures. This is an important difference to recent Graph GPs \citep{ng2018bayesian} and other approaches \citep{venkitaraman2018gaussian} operating in the spectral domain using the graph Laplacian which requires all observations to lie on the same graph (implying the same edge structure).

The dataset $\mathcal{D}=\{\mathcal{G}, \Psi,Y\}$ comprises a collection of signals $\Psi$ with associated labels $Y$, which live on a corresponding domain in $\mathcal{G}$. Let $G \in \mathcal{G}$ be one such domain. This $G = \langle\mathcal{V},\mathcal{E}\rangle$ is comprised of a set of $\mathcal{V}$ vertices and $\mathcal{E} \subseteq \mathcal{V} \times \mathcal{V}$ edges. The corresponding signal $\psi \in \Psi$ is some function $\psi: \mathcal{V} \rightarrow \mathbb{R}^d$ on the vertices of the graph.

We wish to learn some function $f$ which accepts $\psi$ as input and produces an output which will be used for classification, for example passed through a sigmoidal function. It is this function $f$ which we want to model using a Gaussian process prior. Initially, we might view the function of interest as $f: \mathbb{R}^{|\mathcal{V}| \times d} \rightarrow \mathbb{R}$ and select a general covariance function such as a radial basis function (RBF) kernel for the GP prior on $f$. This produces an expressive model, but will suffer from two major drawbacks. First, it completely ignores the underlying graph structures of the inputs, which may contain valuable information for the classification task and may differ substantially between input signals. Second, as $|\mathcal{V}|$ or $d$ increase, a kernel such as the RBF will become more difficult to estimate as it attempts to model the relationships between every input dimension and every vertex, and thus suffers from the curse of dimensionality, a problem common to many expressive models with high-dimensional inputs.

To reduce the complexity of the problem while maintaining expressibility, we will construct kernel functions that focus only on subsets of the input dimensions. These subsets will be defined to take into account the underlying structure of the input graph on which the signal is defined. In particular, we wish to model relationships between nearby vertices while ignoring distant relations, the same intuition that is the basis for convolutional models.

In the context of Gaussian processes, recent work by \citet{convgp} provides a framework for estimating a convolutional GP for classification on Euclidean sampling grids such as images. This work relies fundamentally on additive GP models \citep{duvenaud2011additive11}. The basic insight of these models is that if $\Omega$ defines a set of subsets of the input dimensions, and we model our function of interest as the sum of functions on these subsets $f(x) = \sum_{\omega \in \Omega} g_\omega (\omega)$, where each $g_\omega$ defines some unique function for each $\omega$ modelled using a GP prior $g_\omega \sim \mathcal{GP}(0,k_\omega(\cdot,\cdot))$, then $f$ has an induced GP prior with a covariance function defined by the sum of the respective covariance functions of the constituent $g_\omega$s.

If we focus on the case where $G$ is the same Euclidean sampling grid (image structure) for all observations and constrain $\Omega$ to consist of all $n$-neighborhoods on the interior of the image and $g_\omega$ to be the same $g$ for all $\omega \in \Omega$, we obtain the convolutional Gaussian process model proposed by \citet{convgp}. To be more precise, the subsets in the convolutional GP model when applied to images are easily defined as the $m \times m$ pixel values around each pixel $p$. These are referred to by $\mathbf{x}^{\qb{p}}$ in their paper with $\abs{\mathbf{x}^{\qb{p}}} = m^2$. Image borders can be ignored by only defining subgroups at pixels on the interior of the image, yielding $(w-m+1) \cdot (h - m +1)$ number of pixels $p$ for which there is a $\mathbf{x}^{\qb{p}} \subset \bm{x}$. We can restate this as follows: let the image be represented by the collection of random variables $\bm{x} = \cb{\bm{x}_1, \ldots, \bm{x}_{wh}}$ arranged on an undirected graph with edge weights $\gamma: \mathcal{E} \to \mathbb{R}$ defined as $\gamma(e) = 1$ for all $e\in \mathcal{E}$. We can state the definition of $\mathbf{x}^{\qb{p}}$ as:
\begin{equation}\label{eq:imChunk}
\mathbf{x}^{\qb{p}} = \cb{\bm{x}_j \in \bm{x} \middle| d(\bm{x}_j, \bm{x}_p) \leq n}
\end{equation} 

where $d : \mathcal{V}\times \mathcal{V} \to \mathbb{R}^+$ denotes the distance metric between two vertices defined as:
\begin{equation}\label{eq:graphDistance}
d(\bm{x}_i, \bm{x}_j) = \min_{\pi\in \mathcal{P}} \sum_{e \in \pi} \gamma(e)
\end{equation}
where $\mathcal{P}$ is the set of all paths connecting $\bm{x}_i$ and $\bm{x}_j$ and $e$ is an edge in a given path.  

If $g$ is given a GP prior, a GP prior will be induced on $f$
\begin{equation}
\begin{split}
& g \sim \mathcal{GP}\p{0, k_g\p{\mathbf{t}, \mathbf{t^\prime}}}, \quad f(\mathbf{x}) = \sum_p g\p{\mathbf{x}^{\qb{p}}}, \\
\implies & f\sim \mathcal{GP}\p{0, \sum_{p=1}^P\sum_{p^\prime = 1}^P k_g\p{\mathbf{x}^{\qb{p}}, \mathbf{x}^{\prime\qb{p^\prime}}}}
\end{split}
\end{equation}
where $P$ is the set of pixels and $\mathbf{x}^{\qb{p}}$ indicates the $p^{\text{th}}$ patch of the vector $\mathbf{x}$. (In this literature, what we refer to as $\psi$ is referred to as $\bm{x}$).
Since the subsets are constrained to constitute the same neighborhoods across the image and $g$ is constrained to be the same function on every patch, $g$ becomes a nonlinear, nonparametric patch response function that is translation-invariant. It is this property that mimics the behavior of CNNs.

\section{Graph Convolutional Gaussian Processes}

We now generalize the convolutional GP model to general, non-regular graphs in the spatial domain, so that the model can be applied to a wider range of problems. As mentioned in the previous section, we restrict our focus to a set of graphs in $\mathcal{D}$ that have the same number of vertices. On an arbitrary graph, we can define the same method for grouping the random variables into subsets as set out in Eq. \eqref{eq:imChunk}. The only difference between the settings is that now the underlying graphs do not have the same regular lattice structure as in an image grid. This is a problem because it means that there is no guarantee that $\lvert \mathbf{x}^{\qb{p}}\rvert$ is the same for all $p$, indeed, there is no guarantee that $\abs{\mathbf{x}^{\qb{p}}} = \abs{\mathbf{x}^{\qb{p^\prime}}}$ for \textit{any} $v, v^\prime  \in \mathcal{V}$ for an arbitrary undirected graph $\mathcal{G}$.
An additive GP model could be applied to the set of subsets that are created
\begin{equation}
\begin{split}
g_p \sim \mathcal{GP}\p{0, k_{g_p}\p{\bm{x}_i, \bm{x}_j}} \, \forall p,\\
\quad f(\mathbf{x}) = \sum_p g_p\p{\mathbf{x}^{\qb{p}}}
\end{split}
\end{equation}
We would get a GP prior over $f$, but this would require a separate $g_p$ for every $p$, and would thus not represent a terse, translation-invariant model. Hence, we need to devise a strategy to ensure that $\abs{\mathbf{x}^{\qb{p}}} = \abs{\mathbf{x}^{\qb{p^\prime}}}$ for all $p, p^\prime$. One way of doing this is to throw away all $p$ such that $\abs{\mathbf{x}^{\qb{p}}} \neq c$ for some $c$, but on a general graph this may result in throwing away almost all of the random variables. One could instead find the minimum connectivity of the graph, $m$, and use this to find the $m$-nearest neighbors for each $p$ and then associate these neighbours, along with $p$ itself, to $p$ as $\mathbf{x}^{\qb{p}}$. However, this would also mean throwing away information if the graph is sparsely connected in some areas, but densely connected in others.

To overcome this challenge, we appeal to the literature on graph convolutions (see \citet{niepert2016learning}, \citet{kipf2016semi} or \citet{gdl} for a review of recent work). Spatial-domain charting methods offer the most direct way forward, as they focus on producing a convolution operator, which, when applied to signals on the graph, produces a transformed output with the same dimensionality for every vertex in the graph independent of its connectedness. For a continuous manifold $\mathcal{X}$, the convolution operator can be defined as:
\begin{equation}
\begin{split}
D_j(x)f = \int_{\mathcal{X}} f(x^\prime) u_j(x, x^\prime)\, dx^\prime \\ \forall x \in \mathcal{X}, \quad j = 1\ldots J
\end{split}
\end{equation}
On a discrete graph this becomes:
\begin{equation}\label{eq:discreteconvoperator}
\begin{split}
D_j(v)f = \sum_{\mathcal{V}} f(v^\prime) u_j(v, v^\prime) \\ \forall v \in \mathcal{V}, \quad j = 1\ldots J
\end{split}
\end{equation}
where $u_j$ is some weighting function and $J$ plays the role of the number of bins selected manually. (In this literature, what we refer to as $\psi$ is referred to as $f$). Several possible weighting functions $u_j$ have been proposed in the literature \citep{gdl}. For example, the geodesic polar weighting function \citep{graphNN}, which will be used in subsequent sections for demonstration purposes, is:
\begin{equation}\label{eq:weightFunction}
\begin{split}
u_{j,k}(x, x^\prime) = e^{-\frac{\p{\rho(x^\prime)-\rho_k}^2}{2\sigma_\rho^2}}e^{-\frac{\p{\theta(x^\prime)-\theta_j}^2}{2\sigma_\theta^2}}, \\ j = 1\ldots J, k = 1\ldots K
\end{split}
\end{equation}
where $\rho(\cdot)$ measures the intrinsic radial distance and $\theta(\cdot)$ measures the intrinsic angular distance from $x$ to $x^\prime$. This collection of all $u_{jk}$ forms a system of polar coordinate bins.

\begin{figure}
\centering
\includegraphics[trim=1.7cm 1.36cm 1.22cm 0.9cm, clip,width=.3\columnwidth]{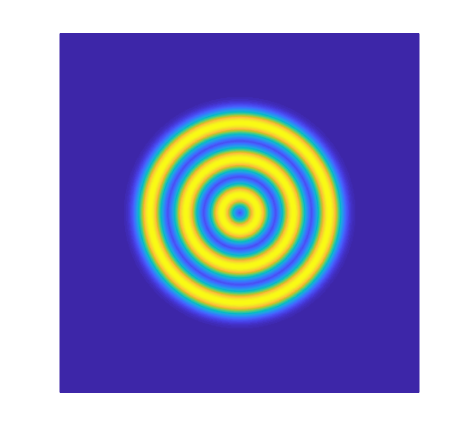}
\includegraphics[trim=2.3cm 1.2cm 1.75cm 0.85cm, clip,height=2.5cm, width=.3\columnwidth]{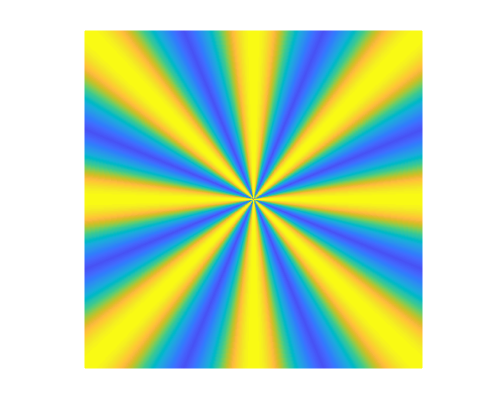}
\includegraphics[trim=2.3cm 1.3cm 1.85cm 0.9cm, clip,width=.3\columnwidth]{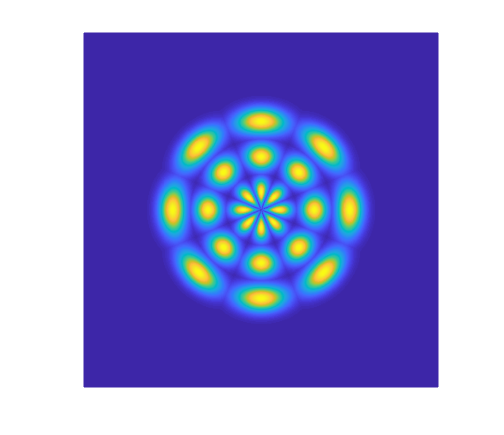}
\caption{(R) The geodesic polar weighting functions in Eq. (\ref{eq:weightFunction}) on an Euclidean sampling grid, with the origin as $x$, and $J=8, K=3$. (L) shows the radial weights, equivalent to the diffusion weighting function. (M) shows the angular weights. }\label{fig:weights}
\vspace{-.5cm}
\end{figure}

There are many potential forms for the weighting functions, however, such as a simpler diffusion function that solely tracks the intrinsic radial distance, anisotropic weighting functions that modify the spread of the weights in certain directions, or general mixtures of Gaussians \citep{moNet}. What these weighting functions all share is the use of some type of coordinates or pseudo-coordinates for the location of the vertices in the graph and some method for finding a distance between these coordinates that is used to create a pairwise importance weighting. How this is implemented is discussed further in the subsequent section.

The GCGP model can be applied to general graphs by appropriate selection of the weighting function $u_j$ found in Eq. \eqref{eq:discreteconvoperator}. Choice and design of this weighting function is one of the main research directions in geometric deep learning and some have been proposed for general graphs. For example, \citet{moNet} use $u_j = e^{-\frac{1}{2} (\mathbf{u}(x,y)-\mathbf{\mu}_j)^\top \mathbf{\Sigma}_j^{-1}(\mathbf{u}(x,y)-\mathbf{\mu}_j)}$ where, for a general graph, one can use pseudo-coordinates $\mathbf{u}(x,y) = \left(\frac{1}{\sqrt{\deg(x)}}, \frac{1}{\sqrt{\deg(y)}}\right)^\top$, where $\deg(\cdot)$ denotes the degree of a vertex. Selecting a different $u_j$ is valid for the underlying machinery of the GP model presented here, provided that the transformation remains linear.

We can now state the GCGP model. Let $\psi, \psi^\prime \in \mathcal{D}$ be $\abs{\mathcal{V}}\times d$ matrices, and let their respective discrete $G, G^\prime$ have $\abs{\mathcal{V}} = \abs{\mathcal{V^\prime}}$. A single $u_{j,k}(x,x')$ is a real value which is the weighting between two vertices, $x$ and $x'$, for a particular set of hyperparameter values, $\rho_k$ and $\theta_j$, of which there are $K$ and $J$ respectively. We then define the tensor $U$ to be $(J \cdot K) \times \abs{\mathcal{V}} \times \abs{\mathcal{V}}$, where each element along the first dimension is a $\abs{\mathcal{V}} \times \abs{\mathcal{V}}$ pairwise weight matrix for a particular set of hyperparameter values. Each element of the weighting matrix is $u_{j,k}$ given some set of hyperparameters.

\begin{figure}[t]
\centering
\includegraphics[width=\linewidth]{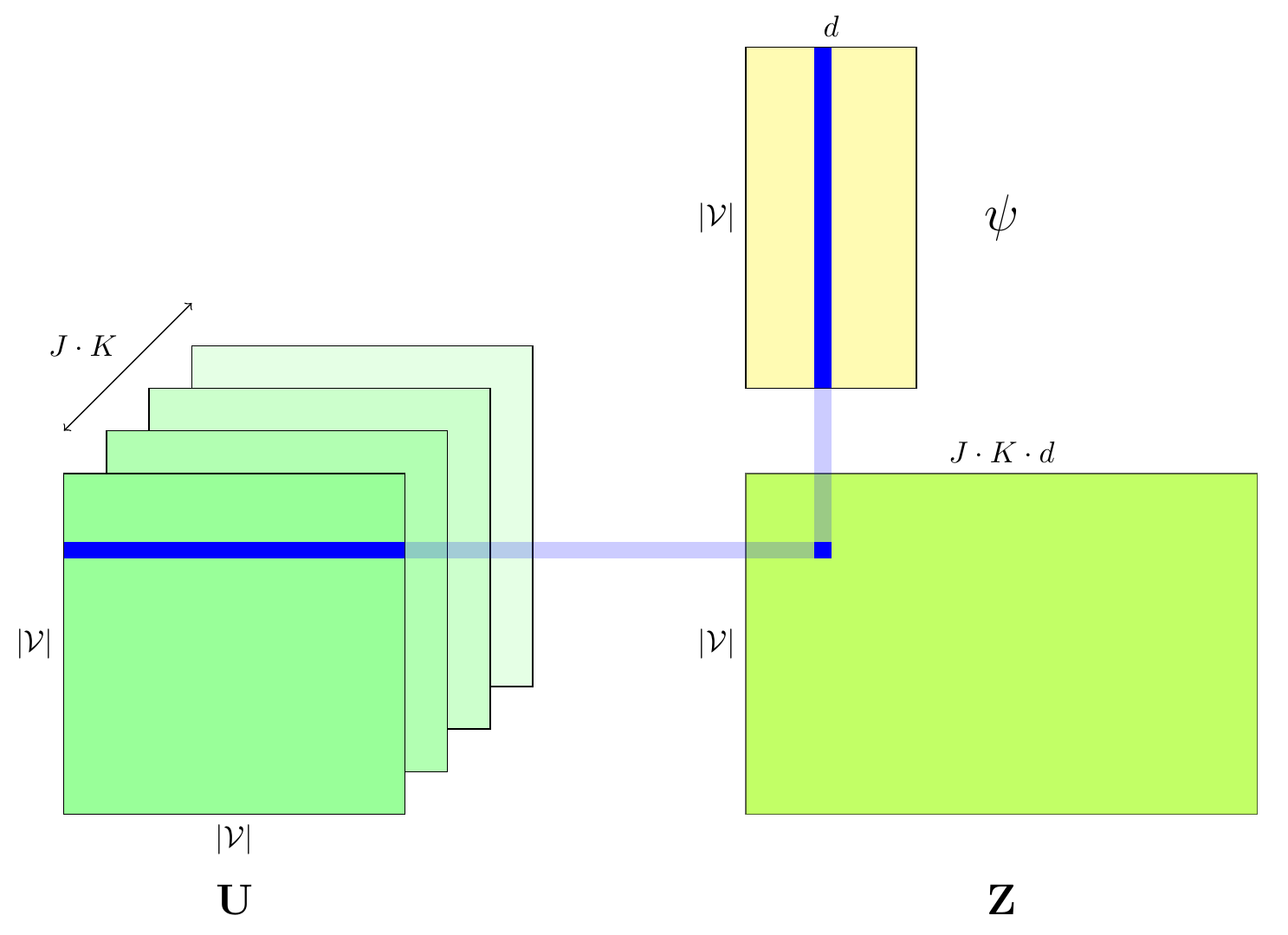}
\caption{Diagram of the construction of patch matrix $Z$ from signal matrix $\psi$ and the weight tensor $U$.}\label{fig:Zmatrix}
\end{figure}

Since the convolution operator is a summation, we can rewrite it as a matrix of inner products we call the convolution patch matrix, $Z$. This will be $\abs{\mathcal{V}} \times (JK \cdot d)$. To get a single element of the $Z$ matrix, we take one element along the first dimension of $U$ ($\abs{\mathcal{V}} \times \abs{\mathcal{V}}$), and then take one row of that ($1 \times \abs{\mathcal{V}}$). From the signal matrix $\psi$ ($\abs{\mathcal{V}} \times d$), we take one column ($\abs{\mathcal{V}} \times 1$). Taking the inner product of these gets us a $1\!\times\!1$, which is a single element of $Z$ (cf. Fig.~\ref{fig:Zmatrix}).


We organize the columns of $Z$ first by dimension of the signal and then by hyperparameter values, so the first $JK$ columns correspond to the first signal dimension, the next $JK$ columns correspond to the second, and so on, appending $d$ $JK$-sized blocks. The size of each column is $\abs{\mathcal{V}}$.

Let $\mathbf{z}^{\qb{i}}$ be the $i^{\text{th}}$ row of the matrix $Z$. This represents the local geodesic polar patch for vertex $i$ of the $d$-dimensional signal $\psi$ as a weighted average along each dimension at each of the $JK$ bin locations.

The function we aim to learn is $f: \mathbb{R}^{\abs{\mathcal{V}} \times d} \to \mathbb{R}$, with input $\psi$, which will be used for classification.  We can think of $f$ as a composition of $h: \mathbb{R}^{\abs{\mathcal{V}}\times (JKd)} \to \mathbb{R}$ and $\mathcal{Z}: \mathbb{R}^{\abs{\mathcal{V}}\times d} \to \mathbb{R}^{\abs{\mathcal{V}}\times (JKd)}$, for which $\psi \mapsto Z$, thus $f(\psi) = h\p{\mathcal{Z}\p{\psi}} = h\p{Z}$. To reduce the dimensionality of the problem and to mimic the translation invariance property we desire, we can now model this $h$ as the summation of a patch response function $g:\mathbb{R}^{JKd} \to \mathbb{R}$, which decomposes as the application of this $g$ to each row of $Z$, i.e. $f(\psi) = h(Z) = \sum_{i \in \mathcal{V}} g(\mathbf{z}^{\qb{i}})$. It is this $g$ that we model using a GP prior with mean zero and some kernel function $k_g$, and thanks to the properties of additive GPs, this will induce a GP prior on $f$:
\begin{equation}\label{eq:graphconvkernel}
\begin{split}
g \sim \mathcal{GP}\p{\mathbf{0}, k_g\p{\mathbf{t}, {\mathbf{t}^\prime}}} \\ \Rightarrow f \sim \mathcal{GP}\p{\mathbf{0}, \sum_{i = 1}^{\abs{\mathcal{V}}}\sum_{j = 1}^{\abs{\mathcal{V^\prime}}} k_g\p{\mathbf{z}^{\qb{i}}, {\mathbf{z}^\prime}^{\qb{j}}}}
\end{split}
\end{equation}
The $k_g$ can be any valid kernel function, but for illustrative purposes an RBF kernel is used. However, this is an interesting choice point for the user since as the number of bins grows and the number of signal dimensions grows, this may suffer from the curse of dimensionality as well, so kernels that further decompose on the patches may become necessary. Since this formulation can be applied to signals on any graphs of the same number of vertices, and $g$ is constrained to be the same patch response function on each extracted local patch, Eq. \eqref{eq:graphconvkernel} can be thought of as a graph convolutional kernel. It is important to note that a similar kernel was proposed for use with support vector machines, but its application was limited to images \citep{convsvm}.

There are several benefits of our formulation. First, while there are numerous potential choices for the weighting functions, the convolution operator itself is linear in the variables that make up the input space. Since the convolution operator is a linear transformation of the collection of random variables, the machinery developed for efficient estimation of the convolutional GP can be applied in our setting as well. This efficient estimation technique relies on the use of interdomain GPs to place inducing points in the space of patches instead of in the input space \citep{interdomainGP}.

Second, the weighting function makes the assumptions about the shape of the convolution on the graph explicit. The standard convolution that we think of on an image is an indicator function that defines membership in a rectangular area the size of which is determined by the investigator. With more general weighting functions, shapes can include rings, ellipses, or more general shapes, making this another interesting choice to explore, as different problems may require different formulations even in the Euclidean domain.

Finally, because the application of the convolution operator is a transformation of the input prior to the application of a valid kernel function, the parameters of the weighting functions $u_{j,k}$ can be treated as hyperparameters of the kernel function. This is because the GCGP forms a manifold GP \citep{manifoldGP}, meaning that the hyperparameters can also be learned at training time, allowing the data to inform the shape of the convolution operator.

\subsection{Estimation of GCGPs}
In the experiments that follow, we estimate the GCGPs using the same estimation technique as used in \citet{convgp}. This method relies on the variational framework for approximation of GPs as proposed in \citet{pmlr-v38-hensman15} which has a computational complexity of $\mathcal{O}(NM^2)$, where $N$ is the number of observations and $M$ is the number of inducing points, and $M$ is chosen such that $M\ll N$.

One may be concerned with the computational complexity of GP methods in general. However, as the overarching motivation is to learn effectively from few training examples, this is not a major practical problem for many interesting applications. Indeed, the often rather small size of datasets that are comprised of objects like meshes is one of the reasons why we think that Bayesian nonparametrics is a promising way forward, even more so when equipped with expressive convolutional feature learning. 

\section{Applications}

In this section, we present applications of GCGPs to both regular domains (images) and non-regular domains (general graphs, meshes). In each application, we use an RBF kernel for the patch response function $g$ and use the polar geodesic weighting function in the construction of the patches. The main challenge in applying GCGPs to different domains is defining the $\rho$ and $\theta$ functions, which measure the radial and angular distance between vertices, respectively. The construction of these will be discussed in more detail in subsequent sections. All experiments were implemented using the GPFlow package \citep{GPflow2017}.

\subsection{MNIST Classification}\label{sec:mnist}

We first consider classification of the standard MNIST dataset in order to illustrate that while the proposed method is general and can handle non-Euclidean domain inputs, it can also be applied to standard datasets with regular sampling grids like images. An image from MNIST is a $28 \times 28$ sampling grid of greyscale pixel values. These pixel values play the role of the signal that lives on the vertices of the graph. We consider the graph structure of an image to be the standard eight-neighbour connectivity, with edge weights equal to 1 everywhere. We define the $\rho$ function in Eq. \eqref{eq:weightFunction} to be the same graph distance as in Eq. \eqref{eq:graphDistance}. We define the $\theta$ function in Eq. \eqref{eq:weightFunction} to be the arctangent of the vertical difference between vertex $v$ and vertex $v'$ over the horizontal difference between the two. Defining these two functions is easy in this setting because we know the orientation of the images, and the orientation remains stable across images. The configuration of the pixels also does not change.

Our main intention here is to compare the performance of our GCGPs to the closest existing Gaussian process model in the literature, which is the convolutional GP model by \citet{convgp}. We follow the same general set-up as used in that paper. In their experiments on the MNIST dataset, they use a $5 \times 5$ patch operator, which produces a 25-dimensional patch response function, modeled using an RBF kernel. To keep comparisons as fair as possible, we use three radial bins and eight angular bins, for a total of a 24-dimensional patch response function. The input signals $\psi: \mathcal{V} \to \mathbb{R}$ are the pixel values that constitute an image. Note that the dimensionality of $\psi$ here is one ($d=1$) as the input is simply a greyscale image. To demonstrate the ability to learn the shape of the convolution, we treat the hyperparameters $\rho_k$ and $\sigma_\rho$ as hyperparameters of the kernel function, while leaving the $\theta_j$ and $\sigma_\theta$ fixed. The $\rho_k$s were initialized to $\{0,1,2\}$, such that the radial bins would be centered on a given pixel, plus the rings one and two pixels away. The $\sigma_\rho$ was initialized to 1, so 68\% of the weight is within one pixel distance from the centeral vertex.

\begin{table}[tb]
	\centering
	\caption{Error rates on MNIST classification}
	\label{tab:MNIST}
	\begin{tabular}{llc}
		\toprule
				&Method & Error rate \\
		\midrule
    \multicolumn{2}{l}{\textbf{MNIST}} & \\ 
		&Conv. GP (25-dim) \dag	& 2.1\% \\
		&RBF GP (784-dim) \dag	& 1.9\% \\
		&\textbf{GCGP (24-dim)}	&\textbf{1.7\%} \\
		\midrule
		\multicolumn{2}{l}{\textbf{MNIST Superpixel 75}} & \\
		& ChebNet \citep{chebnet} & 24.4\% \\
		& MoNet \citep{moNet} & 8.9\% \\
		& \textbf{GCGP} & \textbf{4.2\%}\\
		\midrule
		& \dag\,\citep{convgp} & \\
	\end{tabular}
	\vspace{-0.75cm}
\end{table}

During training, mini-batches of size 200 were used along with 750 inducing points and a learning rate of 0.001. The error rate converges to 1.7\%. This compares favorably to the strictly translation-invariant convolutional GP, which attains an error rate of 2.1\%. An RBF kernel, which models all 784 input dimensions, attains a 1.9\% error rate \citep{convgp}. It is important to note that both the convolutional GP and our GCGP are not hierarchical models, and to increase expressibility, \citet{convgp} include learnable weights which multiply the patch response function for each patch. This increases classification accuracy but means that it is no longer strictly translation-invariant. Our GCGP model can similarly be extended, but this is left for future work. The hyperparameters $\rho_k$ and $\sigma_\rho$ converge to \{1.312, 2.331, 4.034\} and 0.392.

\subsection{MNIST Superpixels}
To demonstrate the performance of GCGPs on general graphs, we apply the model to the 75-vertex MNIST superpixel dataset, following the methodology of \citet{moNet}. This dataset decomposes each MNIST image into a 75-vertex graph with a unique edge structure. This dataset is constructed from the MNIST dataset as follows: for each image, pixels are divided into `background' and `foreground', denoting a pixel value  of 0 and greater than 0 respectively. Then, for each group, k-mean clustering is performed on the pixel locations, the (x,y) index of the pixel, and pixel values. The parameter k is chosen for each group such that their sum is the desired number of superpixels, in this case 75, and so that the split is two thirds foreground and one third background. The resulting centroids give the location of the superpixel in Euclidean coordinates and the value of the superpixel. The value of the superpixels serve as the input signal $\psi: \mathcal{V} \to \mathbb{R}^d$, with $d=1$, which is to be classified. Finally, the graph structure is determined by choosing a threshold value of the Euclidean distance between the centroids and adding an edge if the distance is below the threshold. This produces a unique graph structure for each image. The resulting dataset follows the same training and test set split as the standard MNIST dataset with 60,000 and 10,000 observations in each respectively. The GCGP is set up and trained in precisely the same way as in Section \ref{sec:mnist}. 

The GCGP converges to an error rate of 4.2\% on this task, which is a substantial improvement over the ChebNet architecture proposed by \citet{chebnet} that obtains 24.4\%, and the MoNet architecture that obtains 8.9\% on this task (\citet{moNet}). The improvement against ChebNet is to be expected, as ChebNet is based on learning filters in the spectral domain, which means it will struggle with a task that learns across graphs with different structure, while the GCGP is based on spatial convolutions that can handle such settings. The improvement over the MoNet architecture is more surprising. MoNet can be applied to such tasks and is a deep, hierarchical model in contrast to the GCGP, which is not hierarchical. This may demonstrate the advantage of using such a relatively terse model on a problem of this scale. 

One reason we may prefer a relatively terse Bayesian model to a more complex model in applications to general graphs is we might expect it to perform well even with fewer observations in the training set. To investigate this, we performed an ablation study in which the size of the training set was reduced such that it was composed of 100, 500, or 1000 examples per class, randomly selected from the original training set, which produced new training sets of 1000, 5000, and 10000 examples respectively. The test set was left unchanged.  
The results of this study are presented in Table \ref{tab:ablation}. Remarkably, our GCGP model using only a tenth of the training examples outperforms MoNet trained on the full training set. With only 5,000 training examples (500 per digit) GCGP achieves an error rate of 8.3\% compared to MoNet yielding an error rate of 8.9\% when trained on all 60,000 samples, demonstrating the data efficiency of GCGP. 
\begin{table}[tb]
	\centering
	\caption{Ablation study on Superpixel 75 error rates}
	\label{tab:ablation}
	\begin{tabular}{lccc}
		\toprule
		Examples per class & 100 &  500 & 1000\\
		\midrule
		Error rate & 13.7\% & 8.3\% & 6.3\% \\ 
	\end{tabular}
	\vspace{-.6cm}
\end{table}

\subsection{Triangular 3D Mesh Classification}\label{sec:meshes}


For our next test, we classify triangular meshes. As before, the number of vertices remains the same, but connectivity and embedding in three-dimensional space may be different for each graph.

The data is a collection of 100 meshes from the MPI Faust dataset\footnote{\url{http://faust.is.tue.mpg.de/}} \cite{faust}. This includes ten different poses for each of ten different individuals. We use a training set of 70 meshes where we randomly select 7 poses for each individual. The remaining three poses for each individual comprise the test set (30 observations). Each mesh is comprised of 6800 vertices. These were downsampled to 2500, 1000, and 500 vertices using quadric edge collapse decimation in MeshLab \citep{meshlab} to test the sensitivity of the GCGP to the quality of the meshing.

\begin{figure}
\centering
\includegraphics[width=\linewidth]{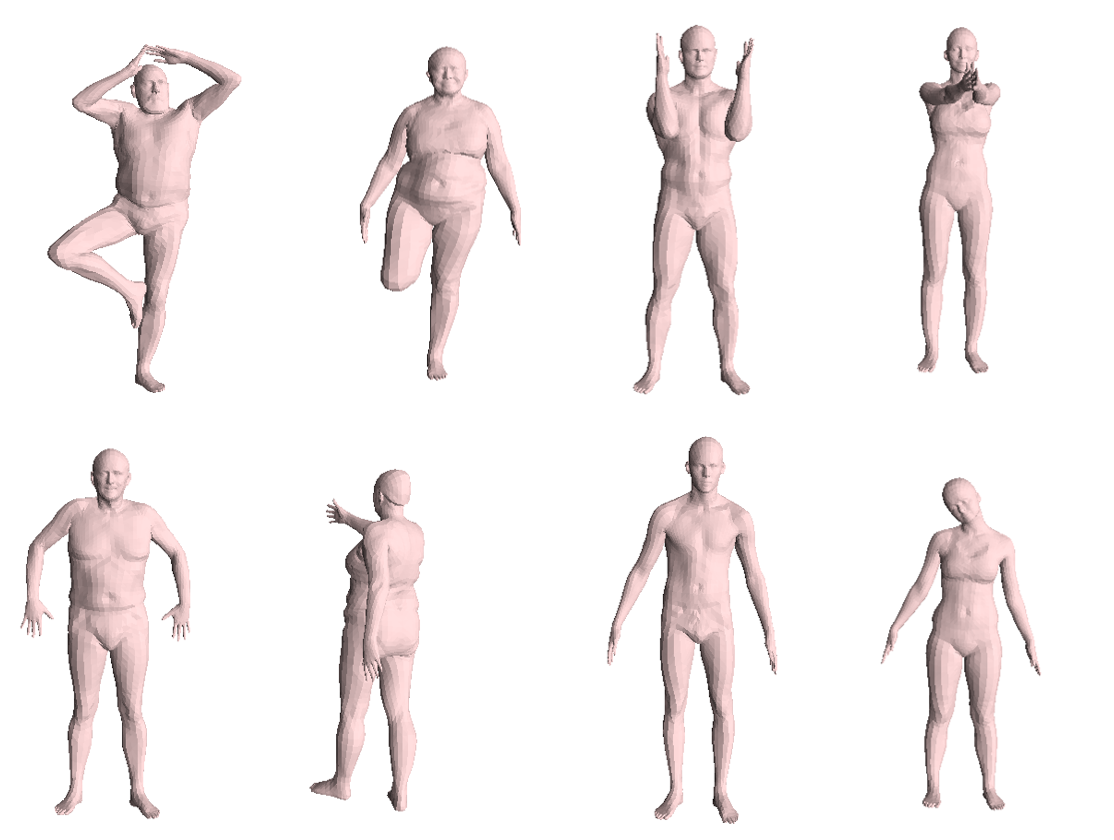}
\caption{Example meshes from the MPI Faust dataset where different people are shown in different poses.}
\end{figure}

\begin{figure}[t]
\centering
\includegraphics[width=\linewidth]{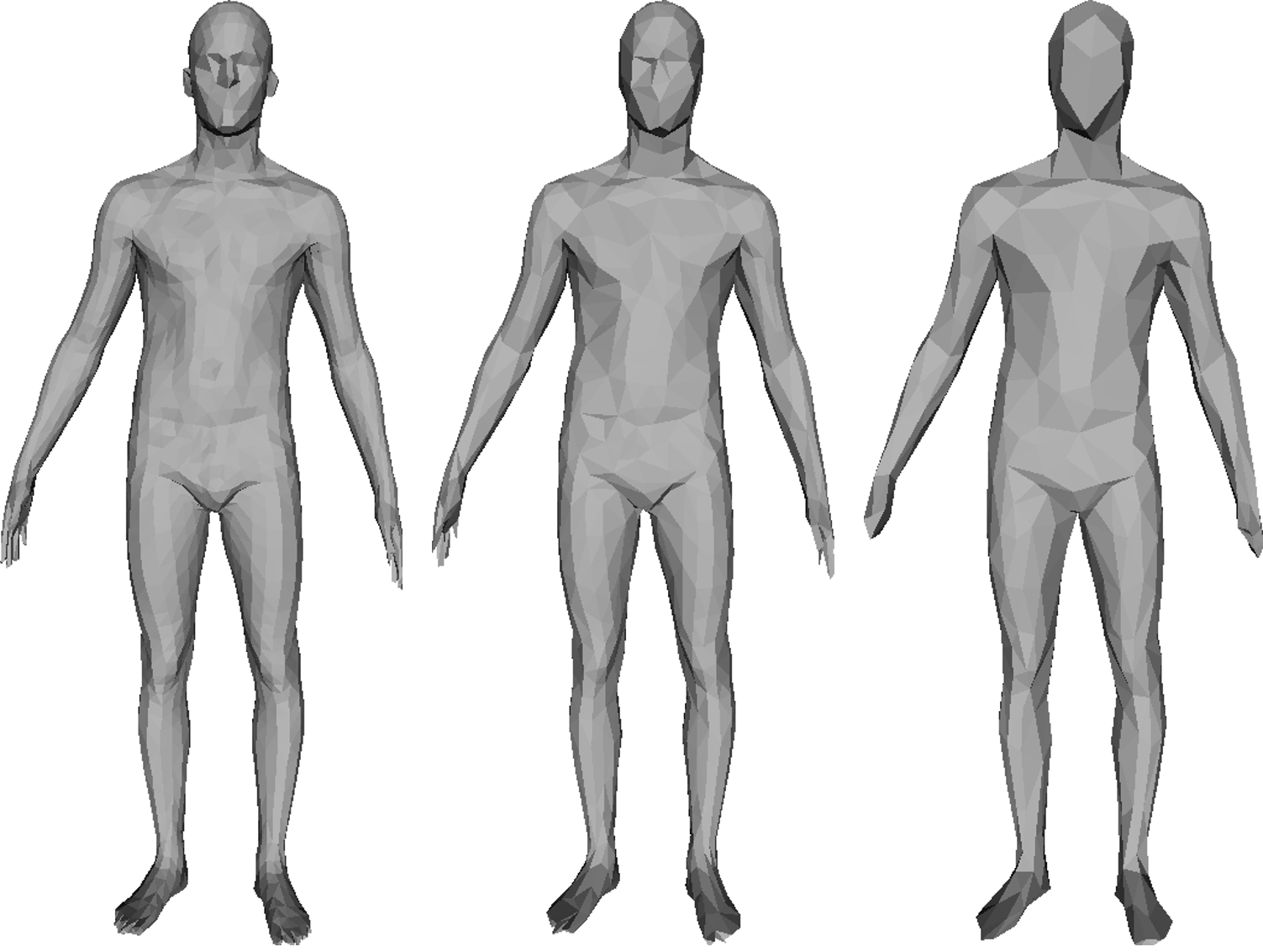}
\caption{MPI Faust meshes resampled to 2500, 1000, and 500 vertices being used in our experiment in Sec. \ref{sec:meshes} to investigate the effect of resolution (cf. Tab. \ref{tab:MPIFAUST}).}
\end{figure}

The goal is to learn the identity of the individual, making this a 10-class classification task. To the best of the authors' knowledge, Gaussian processes have not been employed in a similar task with such types of data.

The input data is a geometric descriptor vector that describes the intrinsic geometry of the mesh, as in \citet{descriptors}. For the purposes of this example, we use only the first four descriptors. For each mesh we have an input signal $\psi: \mathcal{V} \to \mathbb{R}^4$ which is to be classified. We choose 16 angular and 5 radial bins for a total of 80 geodesic polar bins. When combined with the dimensionality of the input, this produces a 320-dimensional patch for each vertex.

We must define the $\rho(\cdot)$ and $\theta(\cdot)$ functions in Eq. \eqref{eq:weightFunction}. This is a more difficult problem than in the Euclidean domain above. We begin by discussing how to compute the $\rho(\cdot)$ function, and subsequently discuss the construction of the $\theta(\cdot)$ function, which is more involved. Any graph pre-processing step in this section was implemented in C++ for efficiency. 

\subsubsection{Radial Distance Between Vertices}

Computing the radial distance between two vertices on a triangular mesh is relatively straightforward, and we follow the same methodology laid out by \citet{graphNN}. We find the intrinsic distance between two vertices on the triangular mesh by using the Fast Marching Method on triangular domains \citep{fmm,fmmGeo}. This method computes the time of arrival for a wave front traveling outwards from a group of initial conditions at a constant rate along the surface of the triangular mesh. By using each vertex as a starting point for the wave front separately, we are able to compute a measure of the intrinsic distance from each vertex to every other vertex, which produces a $\abs{\mathcal{V}} \times \abs{\mathcal{V}}$ matrix of intrinsic distances between the vertices. This can be precomputed for each mesh and does not represent too much of a computational overhead thanks to the $\mathcal{O}(n \log n)$ complexity of the Fast Marching Method. This matrix of distances for each mesh will be used to look up the radial component of the weighting function for each specific pair of vertices in Eq. \eqref{eq:weightFunction}.

\subsubsection{Angular Distance Between Vertices}

To compute $\theta(v,v')$, we must first compute for each vertex $v$ a set of $J$ geodesic rays emanating from $v$. By this we mean a set of rays starting at the central vertex of interest drawn along the surface of the mesh. 


\iftrue 
We begin by taking $J$ rays emanating from a central point and equally dividing $2\pi$ radians between them. We then take these $J$ rays and project them down onto the one-ring of triangles that surround the central vertex $v$. This is done by computing the total angle of the angles that are adjacent to the central vertex, aligning one of the geodesic rays with one of the existing edges of the mesh, and continuing around the central vertex by equally-spaced increments that are proportional to the total angle sum. In this way, each of the $J$ rays will now be a ray lying on one of the faces adjacent to the central vertex. We will now focus on one specific ray and explain how to continue it along adjacent triangles.

We extend the ray to find its intersection point with the edge opposite the central vertex from which it emanates. We now begin a process known as unfolding. This is done by treating the original triangle as lying in a two-dimensional ``unfolding plane''. We then take the triangle from the mesh that shares the edge that the geodesic ray intersects and reconstruct a congruent triangle in the two-dimensional unfolding plane. An alternative way to visualize this is to rotate (around the shared edge) the adjacent triangle in three-dimensional space into the same plane as the starting triangle. We then extend the ray and find the edge of this unfolded triangle with which it intersects. We continue the process of extending the ray along the mesh, unfolding each successive triangle and keeping track of the vertex it passes closest to with each intersection. This process continues until it makes a complete circuit of the mesh, hits the mesh boundary, if one exists, or stops after a designated length.

At this stage, for each of the $J$ geodesic rays, there should be a list of vertices which are the vertices to which the geodesic passed the closest. To create the value of the $\theta$ function from each vertex to every other vertex, we use this list as the initial conditions for Fast Marching and compute the time of arrival for every vertex from this starting point. Once this process is complete, there will be a $J \times \abs{\mathcal{V}} \times \abs{\mathcal{V}}$ tensor that serves as a look-up table for the $\theta$ value for the $j$th angular bin from a given vertex to every other vertex.

If the $\theta_j$ parameters are fixed by the investigator, this can once again be precomputed for each mesh in the dataset. Not fixing the parameters presents an implementation difficulty because if the angular parameters are changed, then it is necessary to recompute all of the geodesics at every update step for the hyperparameters since the $\theta$ parameters control the angles at which the geodesics emanate from the central vertex. Note that this is not a problem for the radial bins since the $\rho$ function, which is simply the intrinsic distances between the vertices on the mesh, remains fixed regardless of the settings of the $\rho$ parameters, which control the midpoint of the radial bin along the surface of the mesh. In the presentation of our results we keep all hyperparameters of the weighting function fixed, which means that we can now use the computed $\rho$ and $\theta$ functions to compute the weight tensors via Eq. \eqref{eq:weightFunction}. Using this, we can transform each signal into a pseudo-image that is $\abs{\mathcal{V}} \times (JK \cdot d)$, which will then be used in the learning task.
\fi

\subsubsection{Results}


\begin{table}[tb]
	\centering
	\caption{Error rates on MPI Faust mesh classification}
	\label{tab:MPIFAUST}
	\begin{tabular}{lccc}
		\toprule
		Number of vertices & 500 &  1000 & 2500\\
		\midrule
		MoNet & 40.00\% & 33.33\% & 33.33\% \\ 
		\textbf{GCGP} & \textbf{23.33\%} & \textbf{10.00\%} & \textbf{3.33\%}\\
	\end{tabular}
	\vspace{-.6cm}
\end{table}

\begin{figure}[t]
\centering
\includegraphics[width=\linewidth]{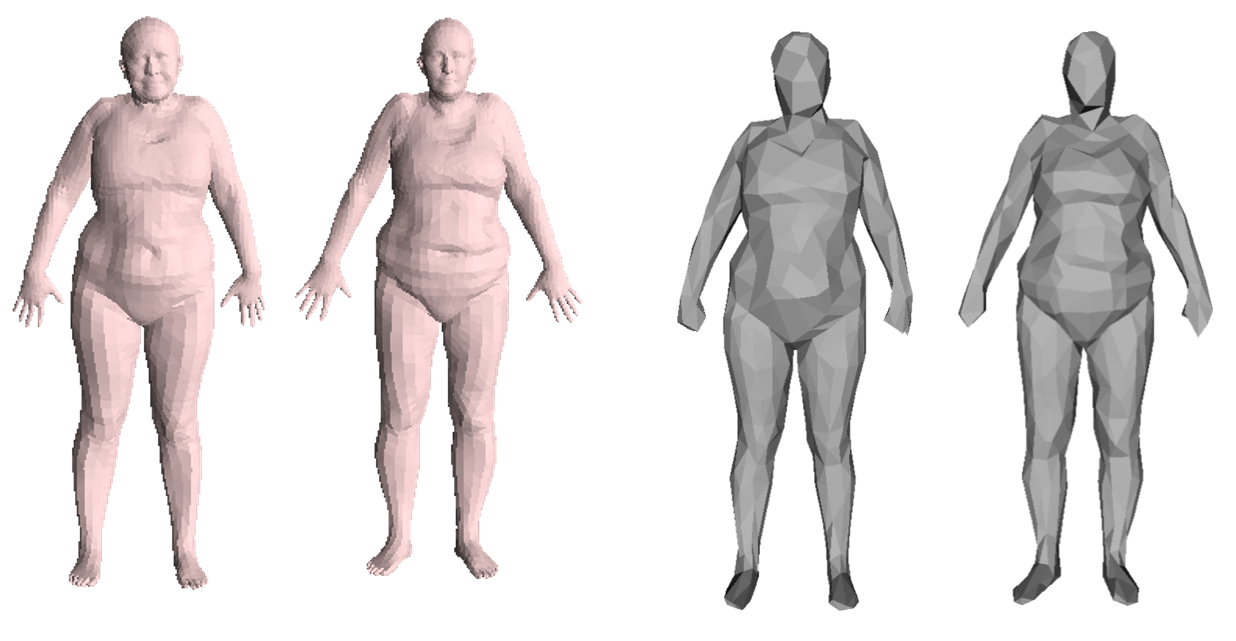}
\caption{Individuals 4 and 8 from MPI Faust with full resolution (left) and 500 vertices (right). GCGP consistently confuses the identity of these two individuals at low resolution, which can be explained by their remarkably high visual similarity.} \label{fig:faust_4_vs_8}
\vspace{-.3cm}
\end{figure}

The results of these experiments are presented in Table \ref{tab:MPIFAUST}. When performing a 10-way classification on the 2500-vertex task, the graph convolutional GP converges to an error of 3.33\%, which represents one misclassification. As expected, reducing the resolution of the mesh decreases accuracy. It is difficult to compare these results to a standard GP model as these must be applied to high-dimensional inputs and few examples per class. Without taking into account any spatial information, these would have 10,000 input dimensions for 2500 vertices, which would be difficult for a standard RBF model to estimate. Indeed, an RBF modeling all 10,000 input dimensions fails to stably learn anything in this case. For these experiments we use a batch size of 30 along with 750 inducing points and a learning rate of 0.001. The results provide some useful, albeit qualitative, insights about the model. In particular, the GCGP model struggles to distinguish between individuals 4 and 8 (Figure \ref{fig:faust_4_vs_8}), especially at the 500-vertex resolution, where it misclassifies every example of one as the other, which can be explained by the high visual similarity of these two subjects.

We also attempted to compare the GCGP performance on the mesh classification task with MoNet as a representative of recent geometric deep learning methods. To this end, we reimplemented MoNet to be able to directly apply it to the 3D mesh data. We initially tried to closely follow the setup as described in \citep{moNet}, i.e., using three graph convolutional layers but with the final layer replaced with a 10 dimensional softmax layer to classify the individual subjects. This model, however, overfitted immediately to the training data of 70 meshes for each of the 10 individuals, yielding a test accuracy close to random guessing. We also explored a less complex MoNet with only one graph convolutional layer for which the best performance we could obtain was an error rate of 33.33\% on the 2500-vertex task. We believe that the small amount of training data is insufficient to properly train this deep learning method, but further investigation into this issue may be required.

\section{Conclusion}

Computer vision and imaging applications have led to significant progress with convolutional methods that excel when data lives on Euclidean domains. The drawback of these methods is that they require large amounts of training data and provide little means of capturing uncertainty when making predictions. Gaussian processes, on the other hand, are an attractive Bayesian approach that can learn from few data points on non-Euclidean domains and provide useful uncertainty estimates. So far, we have seen links between convolutional methods and GPs, leading to convolutional GPs. We have also seen approaches that bring convolutional methods to non-Euclidean domains in the form of graph CNNs, an area also known as geometric deep learning. Here, for the first time, we connect those three areas by combining convolutional methods with GPs on non-Euclidean domains. We believe that graph convolutional GPs provide a powerful new framework enabling exciting new applications and avenues for future research. We believe that the results of the experiments presented here support this, as they demonstrate the GCGP can improve accuracy and make more efficient use of smaller datasets relative to deep learning methods.


We envision augmenting GCGPs to become hierarchical models with recent work on deep GPs \citep{damianou2013deep} as a main directions of future research. This is because the strict translation invariance of the patch response function may not be enough in all situations, which may require combinations of responses across the graph. A simple step is to weigh each response with a set of learnable parameters as in \citep{convgp}.

Another area for improving the model is relaxing the requirement for a fixed number of vertices. In the method presented here, the function that is learned, $f: \mathbb{R}^{\lvert\mathcal{V}\rvert\times d} \to \mathbb{R}$, for classification, of the signals $\psi: \mathcal{V}\to \mathbb{R}^d$ is decomposed in a principled manner into the sum of the responses of a patch response function $g$ across the graph. Modelling $g$ with a GP prior induces a GP prior on $f$, but if $\lvert \mathcal{V} \rvert$ is different for different inputs this setup fails by definition. While there are many situations in which the inputs are signals on graphs with the same number of vertices with potentially different structure, other problems may require the analysis of graphs which have different structure \emph{and} number of vertices, such as classification of molecules. We hope the method presented here can serve as a foundation for GP models that relax this requirement. 

An interesting direction to explore further is the choice of the coordinates/pseudo-coordinates that need not to be the ones used in this paper, and many interesting alternatives exist for meshes and for graphs more generally that are worth investigating. Finally, a nice feature of the GCGP that allows it to be easily tailored to different applications is the selection for the kernel function that describes the patch response function $g$. This is mentioned in the discussion above, but not explored. For example, it may be the case that modelling the patch response function in Section \ref{sec:meshes} with a 320 dimensional RBF kernel is too expressive for the task and the number of observations resulting in suboptimal performance. Performance may therefore be improved by decomposing the patch response function to produce an additive kernel whose components only utilise subsets of the inputs, the same principle that underpins the GCGP.

\section*{Acknowledgements}
This project received funding from the European Research Council (ERC) under the European Union's Horizon 2020 research and innovation programme (grant agreement No 757173, project MIRA, ERC-2017-STG). IW is supported by the Natural Environment Research Council (NERC).


\bibliography{refs}
\bibliographystyle{icml2019}

%
%
%
%

\end{document}